\documentclass{llncs}
\usepackage{graphicx}
\usepackage{subfigure}
\usepackage{cite}
\usepackage{amsmath,amssymb,amsfonts}
\usepackage{algorithmic}
\usepackage{algorithm}
\usepackage{marvosym}
\begin{document}
\title{A Deep Reinforcement Learning Algorithm Using Dynamic Attention Model for Vehicle Routing Problems}
\titlerunning{A Dynamic Attention Model for VRP}
% If the paper title is too long for the running head, you can set
% an abbreviated paper title here
%
\author{ Bo Peng\and
Jiahai Wang\textsuperscript{(\Letter)} \and
%Hong Wu \and
Zizhen Zhang
}
\authorrunning{B. Peng et al.}
% First names are abbreviated in the running head.
% If there are more than two authors, 'et al.' is used.
%
\institute{Department of Computer Science, Sun Yat-sen University, Guangzhou, China \\
\email{wangjiah@mail.sysu.edu.cn}}
\maketitle              % typeset the header of the contribution
\begin{abstract}
Recent researches show that machine learning has the potential to learn better heuristics than the one designed by human for solving combinatorial optimization problems. The deep neural network is used to characterize the input instance for constructing a feasible solution incrementally. Recently, an attention model is proposed to solve routing problems. In this model, the state of an instance is represented by node features that are fixed over time. However, the fact is, the state of an instance is changed according to the decision that the model made at different construction steps, and the node features should be updated correspondingly. Therefore, this paper presents a dynamic attention model with dynamic encoder-decoder architecture, which enables the model to explore node features dynamically and exploit hidden structure information effectively at different construction steps. This paper focuses on a challenging NP-hard problem, vehicle routing problem. The experiments indicate that our model outperforms the previous methods and also shows a good generalization performance.

\keywords{Learning heuristics \and Dynamic encoder-decoder architecture \and Vehicle routing problem \and Reinforcement learning \and Neural network.}
\end{abstract}
\section{Introduction}
Vehicle routing problem (VRP)\cite{c31} is a well-known combinatorial optimization problem in which the objective is to find a set of routes with minimal total costs. For every route, the total demand cannot exceed the capacity of the vehicle. In literature, the algorithms for solving VRP can be divided into exact and heuristic algorithms. The exact algorithms provide optimal guaranteed solutions but are infeasible to tackle large-scale instances due to high computational complexity, while the heuristic algorithms are often fast but without theoretical guarantee. Considering the trade-off between optimality and computational costs, heuristic algorithms can find a suboptimal solution within an acceptable running time for large-scale instances. However, it is non-trivial to design a good heuristic algorithm, since it requires substantial problem-specific expert knowledge and hand-crafted features. Designing a heuristic algorithm is a tedious process, can we learn a heuristic automatically without human intervention?

\begin{figure*}[!t]
\centering
\includegraphics[width=0.7\linewidth]{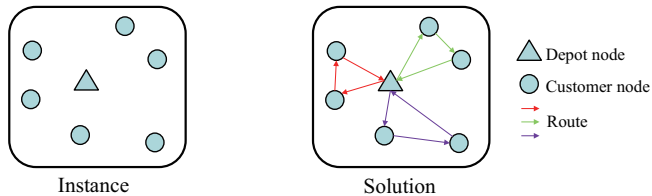}
\caption{A typical vehicle routing problem.}\label{f0}
\end{figure*}

Motivated by recent advancements in machine learning, especially deep learning, there have been some works \cite{c30, c1, c2 ,c3, c4, c5} on using end-to-end neural network to directly learn heuristics from data without any hand-engineered reasoning. Specifically, taking VRP for example, as shown in Fig.~\ref{f0}, the instance is a set of nodes, and the optimal solution is a permutation of these nodes, which can be seen as a sequence of decisions. Therefore, VRP can be viewed as a decision making problem that can be solved by reinforcement learning. From the perspective of reinforcement learning, typically, the state is viewed as the partial solution of instance and the features of each node, the action is the choice of next node to visit, the reward is the negative tour length, and the policy corresponds to heuristic strategy which is parameterized by a neural network. Then the policy is trained to make decisions for maximizing the reward. From the perspective of learning heuristics, given the instances from the distribution $\mathcal{S}$, a heuristics is learned to solve an unseen instance from the same distribution $\mathcal{S}$.

Recently, an attention model (AM) \cite{c5} is proposed to solve routing problems.  In AM, an instance is viewed as a graph, and node features are extracted to represent such a complex graph structure, which captures the properties of a node in the context of its graph neighborhoods. Based on these node features, the solution is constructed incrementally. In AM, the node features are encoded as an embedding which is fixed over time. However, at different construction steps, the state of instance is changed according to the decision the model made, and the node features should be updated correspondingly.

This paper proposes a dynamic attention model (AM-D) with dynamic encoder-decoder architecture. The key of our improvement is to characterize each node dynamically in the context of the graph, which can explore and exploit hidden structure information effectively at different construction steps. To demonstrate the effectiveness of the proposed method, AM-D is applied to a challenging combinatorial optimization problem, vehicle routing problem. The numerical experiments indicate that AM-D performs significantly better than AM and obviously decreases the optimality gap.

This paper is structured as follows. Section \ref{s2} reviews related work. Section \ref{s4} discribes original attention model for VRP. Section \ref{s5} and \ref{s6} present our dynamic attention model for VRP and the training method, respectively. Experimental results are given in Section \ref{s7}. Section \ref{s8} concludes this paper.  \label{s1}

\section{Related Work}\label{s2}
Learning heuristic based methods proposed in last serval years can be divided into two categories in terms of types of problems solved. The first category focuses on solving permutation based combinatorial optimization problems, such as VRP and TSP. The second category solves 0-1 based combinatorial optimization problems, such as SAT and knapsack problem.

For the first category, the pointer network (PN) is introduced in \cite{c8}, it takes combinatorial optimization problems as sequence to sequence problems where the input is a sequence of nodes and the output is a permutation of the input. PN overcomes the limitation that the output length depends on input by a ``pointer", which is a variant of attention mechanism \cite{c6}. This sequence to sequence model \cite{c11} is trained by the supervised manner and the label is given by an approximate solver.

However, PN is sensitive to the quality of labels and optimal solutions are expensive. In \cite{c1}, the neural combinatorial optimization framework is proposed to solve combinatorial optimization problems, and the REINFORCE algorithm \cite{c9} is used to train a policy modeled by PN without supervised signals. In\cite{c2}, the LSTM encoder of PN is replaced by element-wise projections which are invariant to the input order and will not introduce redundant sequential information.

In \cite{c3}, combinatorial optimization is taken as a graph problem, and graph embedding \cite{c12} is used to capture combinatorial structure information between nodes. The model is trained by 1-step DQN \cite{c13} which is data-efficient, and the solution is constructed by the helper function.

In \cite{c4} and \cite{c5}, graph attention network \cite{c7} is used to extract the features of each node in graph structure. In\cite{c4}, an explicitly forgetting mechanism is introduced to construct a solution, which only requires the last three selected nodes per step. Then the constructed solution is improved by 2OPT local search \cite{c14}. In \cite{c5}, a context vector is introduced to represent the decoding context, and the model is trained by the REINFORCE algorithm with a deterministic greedy rollout baseline.

For the second category, in \cite{c16}, the graph convolutional network \cite{c15, c17} is trained to estimate the likelihood, for each node in the instance, of whether this node is part of the optimal solution. In addition, the tree search is used to construct a large number of candidate solutions. In \cite{c33}, GCOMB is proposed to solve combinatorial optimization problems over large graph based on graph convolutional network and Q-learning. In \cite{c18} and \cite{c19}, the model is taken as a classifier. In \cite{c18}, the message passing neural network \cite{c20} is trained to predict satisfiability on SAT problems. In \cite{c19}, the graph neural network is used to solve decision TSP.

Since this study is targeted at solving VRP, \cite{c2,c5} are the most related work with this paper. AM proposed recently in \cite{c5} for VRP is introduced as follows.

\begin{figure*}[t]
\centering
\subfigure[]{
\includegraphics[width=0.9\linewidth]{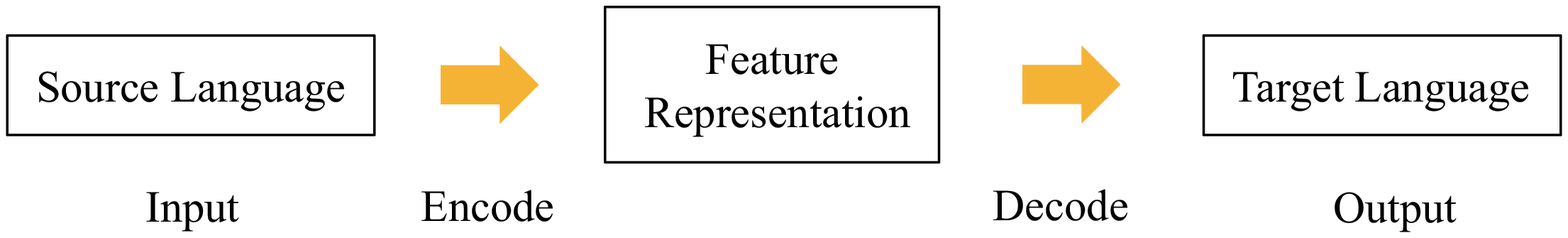}\label{f2a}}
\subfigure[]{
\includegraphics[width=0.9\linewidth]{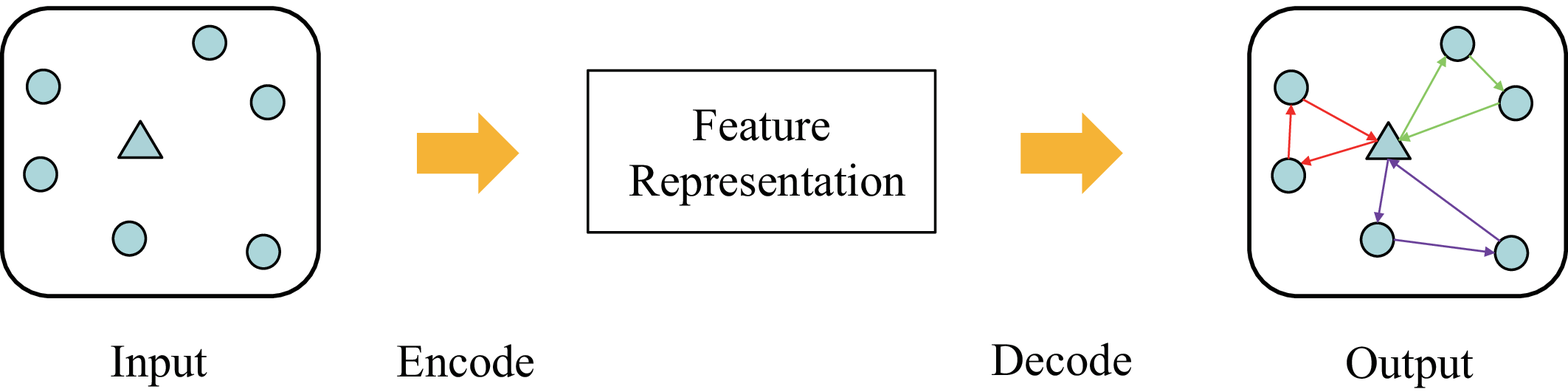}\label{f2b}}
\caption{(a) The encoder-decoder architecture for NMT. (b) The encoder-decoder architecture for VRP.}\label{f2}
\end{figure*}

\section{Attention Model for VRP}\label{s4}

\subsection{Problem Formulation and Preliminaries}\label{s3}

This paper focuses on VRP. For the simplest form of the VRP, a single capacitated vehicle is responsible for delivering items to multiple customer nodes, and the vehicle must return to the depot to pick up additional items when it runs out of loads. The solution can be seen as a set of routes. In each route, it begins and ends at the depot.

Specifically, for VRP instance, the input $X=\{x_0,\ldots, x_n\}$ is a set of nodes and $x_0$ is the depot. Each node consists of two elements $x_i=(s_i,d_i)$, where $s_i$ is a 2-dimensional coordinate of node $i$ in euclidean space and $d_i$ is its demand ($d_0=0$). The solution $\pi$ is a sequence $\{\pi=(\pi_1,\ldots, \pi_T) , \pi_t\in\{x_0,\ldots,x_n\}\}$, where each customer node is visited exactly once and the depot can be visited multiple times. $T$ is the length of sequence $\pi$ that may be varied from different solutions.

VRP can be viewed as a sequential decision making problem, and encoder-decoder architecture \cite{c11} is an effective framework for solving such kind of problems. Taking neural machine translation (NMT) for example, as shown in Fig.~\ref{f2a}, the encoder extracts syntactic structure and semantic information from source language text. Then the decoder constructs target language text from the features given by encoder. Fig.~\ref{f2b} shows that the encoder-decoder architecture can also be applied to solve VRP. Firstly, the structural features of the input instance are extracted by the encoder. Then the solution is constructed incrementally by the decoder. Specifically, at each construction step, the decoder predicts a distribution over nodes, then one node is selected and appended to the end of the partial solution. Hence, corresponding to the parameters $\theta$ and input instance $X$, the probability of solution $p_\theta(\pi\vert X)$ can be decomposed by chain rule as:

\begin{equation}
p_\theta(\pi\vert X)=\prod_{t=1}^Tp_\theta(\pi_t|	X,\pi_{1:t-1}). \label{eq1}
\end{equation}

\subsection{Encoder}

In encoder, graph attention network is used to encode node features to an embedding in context of graph. It is similar to the encoder in transformer architecture \cite{c21}. Firstly, for each $d_x$-dimensional (for VRP, $d_x=3$, the coordinate and demand) input node $x_i$, the $d_h$-dimensional ($d_h=128$) initial node embedding $h_i^{(0)}$ is computed through a linear transformation with learnable parameters $W\in\mathbb{R}^{d_h\times d_x}$ and $b\in\mathbb{R}^{d_h}$, separate parameters $W_0$ and $b_0$ are used for the depot:

\begin{equation}
h_i^{(0)}=
\begin{cases}
Wx_i+b & \mbox{if } i\ne0\\
W_0x_i+b_0 & \mbox{if } i=0.\\
\end{cases}\label{eq2}
\end{equation}

These initial node embeddings are fed into the first layer of graph attention network and updated $N=3$ times with $N$ attention layers. For each layer, it consists of two sublayers: a multi-head attention (MHA) sublayer and a fully connected feed-forward (FF) sublayer.

\subsubsection{Multi-Head Attention Sublayer}
As in \cite{c21}, multi-head attention is used to extract different types of information. In the layer $\ell\in \{1,\ldots,N\}$, $h_i^{(\ell)}$ is denoted as the node embedding of each node $i$, and the output $\{h_0^{(\ell-1)}, \ldots,h_n^{(\ell-1)}\}$ of the layer $\ell-1$  is the input of the layer $\ell$. The multi-head attention vector $\text{MHA}_i^{(\ell)}(h_0^{(\ell-1)},\ldots,h_n^{(\ell-1)})$ of each node $i$ can be computed as:

 \begin{equation}
 q_{im}^{(\ell)}=W_{m}^Qh_{i}^{(\ell-1)},k_{im}^{(\ell)}=W_{m}^Kh_{i}^{(\ell-1)}, v_{im}^{(\ell)}=W_{m}^Vh_{i}^{(\ell-1)},\label{eq3}
\end{equation}

\begin{equation}
 u_{ijm}^{(\ell)}=(q_{im}^{(\ell)})^Tk_{jm}^{(\ell)},\label{eq4}
\end{equation}

\begin{equation}
 a_{ijm}^{(\ell)}=\frac{e^{u_{ijm}^{(\ell)}}}{\sum_{j^{'}=0}^{n}e^{u_{ij^{'}m}^{(\ell)}}},\label{eq5}
\end{equation}

 \begin{equation}
 h_{im}^{'(\ell)}=\sum_{j=0}^{n}a_{ijm}^{(\ell)}v_{jm}^{(\ell)},\label{eq6}
\end{equation}

\begin{equation}
 \text{MHA}_i^{(\ell)}(h_0^{(\ell-1)},\ldots,h_n^{(\ell-1)})=\sum_{m=1}^MW_m^Oh_{im}^{'(\ell)}.\label{eq7}
\end{equation}

Here, the number of head is set $M=8$, in each attention head $m\in\{1,\ldots\,M\}$, the query vector $q_{im}^{(\ell)}\in\mathbb{R}^{d_k}$, key vector $k_{im}^{(\ell)}\in\mathbb{R}^{d_k}$ and value vector $v_{im}^{(\ell)}\in\mathbb{R}^{d_v}$ are computed with parameters $W_{m}^Q\in\mathbb{R}^{d_k\times d_h}$, $W_{m}^K\in\mathbb{R}^{d_k\times d_h}$ and $W_{m}^V\in\mathbb{R}^{d_v\times d_h}$ respectively. And the final vector is computed with $W_{m}^O\in\mathbb{R}^{d_h\times d_v}$ ($d_k=d_v=\frac{d_h}{M}=16$).

\textbf{Remark}: the parameters $W_{m}^Q$, $W_{m}^K$ and $W_{m}^V$ do not share between each layer and the superscript $\ell$ is omitted for readability.

\subsubsection{Feed-Forward Sublayer}
In this sublayer, for each node $i$, based on multi-head attention vector, $h_i^{(\ell)}$  is computed by skip-connection and fully connected feed-forward (FF) network. For each node $i$:

\begin{equation}
 \hat{h}_i^{(\ell)}=\tanh(h_i^{(\ell-1)}+\text{MHA}_i^{(\ell)}(h_0^{(\ell-1)},\ldots,h_n^{(\ell-1)})),\label{eq8}
\end{equation}

\begin{equation}
 \text{FF}(\hat{h}_i^{(\ell)})=W_{1}^{F} \text{ReLu}(W_{0}^{F}\hat{h}_i^{(\ell)}+b_{0}^{F})+b_{1}^{F},\label{eq9}
\end{equation}

\begin{equation}
 h_i^{(\ell)}=\tanh(\hat{h}_i^{(\ell)}+\text{FF}(\hat{h}_i^{(\ell)})),\label{eq10}
\end{equation}
where $h_i^{(\ell)}$ is calculated with parameters $W_{0}^{F}\in\mathbb{R}^{d_F\times d_h} $, $W_{1}^{F}\in\mathbb{R}^{d_h\times d_F}$, $b_0^{F}\in\mathbb{R}^{d_F}$ and $b_1^{F}\in\mathbb{R}^{d_h} $($d_F=4\times d_h$).

After $N$ attention layers, for each node $i$, the final node embedding $h_i^N$ is calculated as:
\begin{equation}
 h_i^{N}=\text{ENCODE}_i^{N}(h_0^0,\dots,h_n^0).\label{eq11}
\end{equation}
$\text{ENCODE}^{N}_i(h_0^0,\dots,h_n^0)$ is computed with Eqs.~\eqref{eq3}-\eqref{eq10}.

\begin{figure}[t]
\centering
\includegraphics[width=0.5\linewidth]{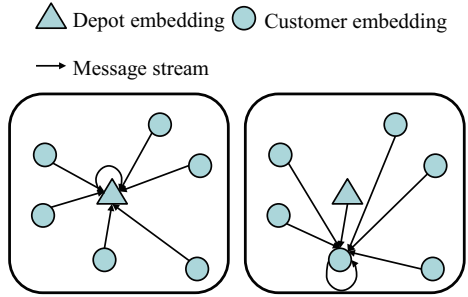}
\caption{The message stream in encoder. The embedding of each node is updated by aggregating the message of each node (including itself). On the left, the depot embedding is updated. On the right, the customer embedding is updated. }\label{f3}
\end{figure}

Fig.~\ref{f3} illustrates the stream of message between nodes. By aggregating the message of each node, the embedding of each node is updated according to the attention mechanism.

\subsection{Decoder}
In decoder, at each construction step $t\in\{1,\ldots,T\}$, one node is selected to visit based on the partial solution $\pi_{1:t-1}$ and the embedding of each node. As in \cite{c5}, the context vector $h_c$ is computed by $M$-head attention mechanism. Firstly, for VRP, a new vector $h_c^{'}$ is constructed as:
\begin{equation}
h_c^{'}=
\begin{cases}
[\bar{h}_t;h_0^N;D_t] & \mbox{if } t=1\\
[\bar{h}_t;h_{\pi_{t-1}}^N;D_t] & \mbox{if } t>1,\\
\end{cases}\label{eq12}
\end{equation}
where [ ; ] is concatenation operator, $h_{\pi_{t-1}}^N$ is the embedding of the node selected at construction step $t-1$, $D_t$ is the remaining capacity of vehicle ($D_1=D$), and $\bar{h}_t$ is the graph embedding, which is the mean vector of the embedding over nodes that have not been visited (including depot) at construction step $t$. Similar to the encoder, $h_c$ is computed with a single $M$-head attention layer, and only a single query $q_{(c)}$ (per head) is computed (the parameters do not share with encoder):

 \begin{equation}
 q_{(c)m}=W_{m}^Qh_c^{'},\quad k_{jm}=W_{m}^Kh_j^N,\quad v_{jm}=W_{m}^Vh_j^N,\label{eq13}
\end{equation}

\begin{equation}
u_{(c)jm}=
\begin{cases}
q_{(c)m}^Tk_{jm} & \mbox{if } d_j<=D_t \text{ and }x_j\notin\pi_{1:t-1}\\
-\infty &  \text{otherwise}, \\
\end{cases}\label{eq14}
\end{equation}

\begin{figure*}[t]
\centering
\includegraphics[width=1\linewidth]{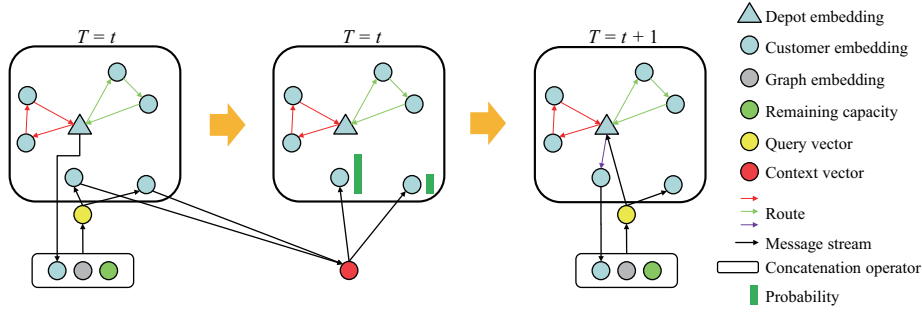}
\caption{The details of decoder at construction step $t$. At each construction step, according to the context vector and node embedding (except the nodes that violates the constraints), the decoder predicts a distribution over nodes and selects one to visit.}\label{f4}
\end{figure*}

\begin{equation}
 a_{(c)jm}=\frac{e^{u_{(c)jm}}}{\sum_{j^{'}=0}^{n}e^{u_{(c)j^{'}m}}},\label{eq15}
\end{equation}

 \begin{equation}
 h_{(c)m}^{'}=\sum_{j=0}^{n}a_{(c)jm}v_{jm},\label{eq16}
\end{equation}

\begin{equation}
  h_{c}=\sum_{m=1}^MW_m^Oh_{(c)m}^{'}.\label{eq17}
\end{equation}

As shown in Eq.~\eqref{eq14}, in order to construct a feasible solution, the node that violates the constraints will be masked. For VRP, the following masking conditions are used. First, the customer node whose demand greater than the remaining capacity of the vehicle is masked. Second, the customer node that already been visited is masked.

\textbf{Remark}: the depot node can be visited multiple times and it will be masked only when $\pi_{t-1}=x_0$.

Finally, the probability $p_\theta(\pi_t|X,\pi_{1:t-1})$ is computed with a single-head attention layer:

 \begin{equation}
 q=W^Qh_c,\quad k_j=W^Kh_j^N,\label{eq18}
\end{equation}

\begin{equation}
u_{j}=
\begin{cases}
C\cdot \tanh (q^Tk_j) & \mbox{if } d_j<=D_t \text{ and }x_j\notin\pi_{1:t-1}\\
-\infty &  \text{otherwise}, \\
\end{cases}\label{eq19}
\end{equation}

\begin{equation}
 p_\theta(\pi_t=x_j|X,\pi_{1:t-1})=\frac{e^{u_{j}}}{\sum_{j^{'}=0}^{n}e^{u_{j^{'}}}},\label{eq20}
\end{equation}
where $C$ is used to clip the result within $[-C, C]$ ($C=10$). If node $i$ is selected to visit at construction step $t$, the remaining capacity should be updated:
\begin{equation}
D_{t+1}=
\begin{cases}
D & \mbox{if } i=0\\
D_t-d_i &  \text{otherwise}. \\
\end{cases}\label{eq21}
\end{equation}

Fig.~\ref{f4} illustrates the details of decoder at construction step $t$. According to the partial solution and node embedding, the context vector is computed by the attention mechanism. Based on the context vector and the embedding of remaining nodes, the decoder predicts a distribution over these nodes and selects one to visit.

\section{Dynamic Attention Model for VRP}\label{s5}
\begin{figure*}[!t]
\centering
\includegraphics[width=1\linewidth]{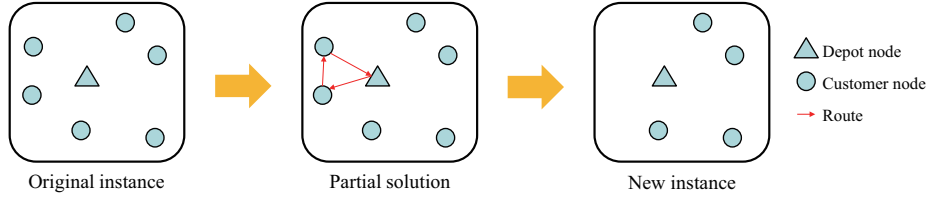}
\caption{The state of an instance is changed at different construction steps. When the model constructed a partial solution, the remaining nodes can be seen as a new instance. Since some nodes already been visited, the structure of this new instance is different from the original instance. }\label{f1}
\end{figure*}

\begin{figure*}[!b]
\centering
\subfigure[]{
\includegraphics[width=0.4\linewidth]{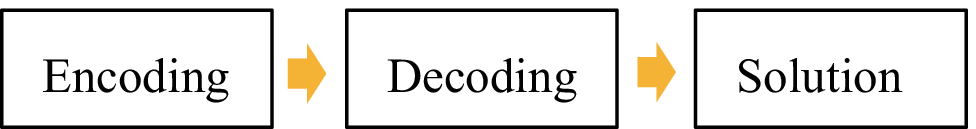}\label{f5a}}
\subfigure[]{
\includegraphics[width=1\linewidth]{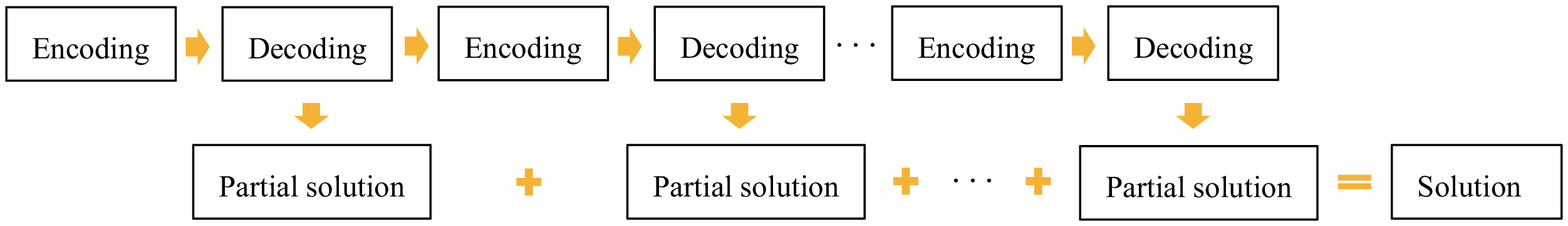}\label{f5b}}
\caption{The comparison between our dynamic architecture and the vanilla architecture. (a) In vanilla encoder-decoder architecture of AM for VRP, the encoder is used only once, the embedding of each node is fixed, which only can represent the initial state of the input instance. (b) In dynamic encoder-decoder architecture of AM-D, the encoder and decoder are used alternately to recode the embedding of each node and construct a partial solution.}\label{f5}
\end{figure*}

As mentioned in Section \ref{s4}, the solution is constructed incrementally by the decoder. At different construction steps, the state of the instance is changed, and the feature embedding of each node should be updated. As shown in Fig.~\ref{f1}, when the model constructed a partial solution, the remaining nodes, which do not be included in the partial solution yet, can be seen as a new instance. Constructing the remaining solution is equivalent to solve this new instance. Since some nodes have already been visited, the structure of this new instance is different from the original instance. Therefore, the structure information is changed and the node features should be updated accordingly. But in vanilla encoder-decoder architecture in AM for VRP, as shown in Fig.~\ref{f5a}, the feature embedding of each node is computed only once, which corresponds to the initial state of instance. This paper proposes a dynamic encoder-decoder architecture to characterize the feature embedding of each node dynamically at different construction steps.

The dynamic encoder-decoder architecture, as shown in Fig.~\ref{f5b}, is similar to vanilla encoder-decoder architecture. The key difference is that the embedding of each node will be immediately recomputed when the vehicle returns to the depot. Specifically, for each node $i$, the embedding can be updated at construction step $t$ as:
\begin{equation}
h_{i}^{t}=
\begin{cases}
\text{ENCODE}^{N}_i(h_0^0,\dots,h_n^0) & \mbox{if } \pi_{t-1}=x_0\\
h_{i}^{t-1} &  \text{otherwise}, \\
\end{cases}\label{eq22}
\end{equation}
where $h_i^t$ is the embedding of node $i$ at construction step $t$, and the layer number $N$ is omitted. $\text{ENCODE}^{N}_i(h_0^0,\dots,h_n^0)$ is similar to  Eq.~\eqref{eq11} that is computed with $N$ $M$-head attention layers. The only difference is that  Eq.~\eqref{eq4} is modified. In order to reflect that the structure of instance is changed, the nodes that have been visited are masked, and  Eq.~\eqref{eq4} is modified as:

\begin{equation}
u_{ijm}^{(\ell)}=
\begin{cases}
(q_{im}^{(\ell)})^Tk_{jm}^{(\ell)} & \mbox{if } x_j\notin\pi_{1:t-1} \text{ or }j=0\\
-\infty &  \text{otherwise}. \\
\end{cases}\label{eq23}
\end{equation}

During decoding, at each step $t$, the computation of  Eqs.~\eqref{eq13}-\eqref{eq18} is based on the latest embedding of each node $\{h_0^t,\dots,h_n^t\}$ (the layer number $N$ is omitted, and $t$ is the construction step). As shown in Fig.~\ref{f5b}, the entire architecture uses the encoder and decoder alternately to recode node embedding and construct a partial solution.

Given a distribution over nodes, there are two strategies to select the next node to visit. The one is sample rollout that selects a node using sampling. The other is greedy rollout that selects the node with maximum probability. The former is a stochastic policy and the latter is a deterministic policy.

\section{Model Training}\label{s6}
As in \cite{c1, c2, c4, c5}, solving combinatorial optimization problem is taken as Markov Decision Processes (MDP), and AM-D is trained by policy gradient using REINFORCE algorithm\cite{c9}. Given an instance $X$, our training objective is the tour length of solution $\pi$. Hence, based on instance $X$, the gradients of parameters $\theta$ are defined as:

\begin{equation}
\nabla_{\theta}J(\theta | X) =\mathbb{E}_{\pi \sim p_{\theta}(.|X)}[(L(\pi|X)-b(X))\nabla_{\theta}\log p_{\theta}(\pi|X)],\label{eq26}
\end{equation}
where $L(\pi|X)$ is the tour length of solution $\pi$, $b(X)$ is a baseline function for estimating the expected tour length $\mathbb{E}_{\pi \sim p_{\theta}(.|X)}L(\pi|X)$ of instance $X$ which can reduce the variance of gradients and accelerate convergence effectively. In this paper, as in \cite{c5}, the tour length of the greedy solution, which is constructed by greedy rollout, is taken as $b(X)$.

During training, the instances are drawn from the same distribution $\mathcal{S}$. The gradients of parameters $\theta$ are approximated by Monte Carlo sampling as:

\begin{equation}
\nabla_{\theta}J(\theta) \approx \frac{1}{B}\sum_{i=1}^B [(L(\pi_i^{s}|X_i)-L(\pi_i^{g}|X_i))\nabla_{\theta}\log p_{\theta}(\pi_i^s|X_i)],\label{eq27}
\end{equation}
where $B$ is the batch size, $\pi_i^s$ and $\pi_i^g$ are the solutions of instance $X_i$ constructed by sample rollout and greedy rollout respectively. The training algorithm is described in Algorithm \ref{a1}.

\begin{algorithm}[H]
  \centering
  \scriptsize
  \caption{REINFORCE algorithm}
  \begin{algorithmic}[1]
  	  \STATE {\bfseries Input:} number of epochs $E$, steps per epoch $F$, batch size $B$
      \STATE Initialize parameters $\theta$
      \FOR{$\text{epoch} = 1, \ldots, E$}
        \FOR{$\text{step} = 1, \ldots, F$}
        	\STATE $X_i \gets \text{RandomInstance()} \enspace \text{for}\enspace i \in \{1, \ldots, B\}$
            \STATE $\pi_i^{s} \gets \text{SampleRollout}(p_{\theta}(.|X_i)) \enspace  \text{for}\enspace i \in \{1, \ldots, B\}$
            \STATE $\pi_i^{g} \gets \text{GreedyRollout}(p_{\theta}(.|X_i)) \enspace  \text{for}\enspace i \in \{1, \ldots, B\}$
            \STATE $g_{\theta} \gets \frac{1}{B}\sum_{i=1}^{B} \left(L(\pi_i^s) - L(\pi_i^g)\right) \nabla_{\theta} \log p_{\theta}(\pi_i^s|X_i)$

          	\STATE $\theta \gets \text{Adam}(\theta, g_{\theta})$
        \ENDFOR
     \ENDFOR
  \end{algorithmic}\label{a1}
\end{algorithm}

\section{Experiments}\label{s7}
Experiments are conducted to investigate the performance of AM-D on VRP with node size $n=20,50,100$. AM-D consists of two phases: training phase and testing phase. For each problem, in training phase, the model is trained with 30 epochs, and 10000 batches are processed in each epoch. In testing phase, the performance on 10000 test instances is reported, where the solution is constructed by greedy rollout, and the final results are the average length on all test instances.

\subsection{Instances and Hyperparameters}

As in \cite{c2} and \cite{c5}, the instances are generated from a fixed distribution. For each node, the location are chosen randomly from the unit square $[0,1]\times[0,1]$, and the demand is a discrete number in $\{1,\dots,9\}$ chosen uniformly at random (the demand of depot is $0$). The capacity of vehicle $D=30$ for VRP with 20 customer nodes (denoted as VRP20), $D=40$ for VRP50, $D=50$ for VRP100, and the vehicle is located at the depot when $t=1$. The batch size $B=128$ and learning rate $\eta=10^{-4}$ for both VRP20 and VRP50, $B=108$ and $\eta=5\times10^{-5}$ for VRP100. Finally, for each problem, the experiment is conducted by GPU (single 1080Ti for VRP20, VRP50, 3$\times$1080Ti for VRP100).

\begin{table}[t]
\caption{Results on VRP. Len is the average length on test instance. Gap is the distance to state-of-the-art.}
\centering
\renewcommand\arraystretch{1.25}
\begin{tabular}{| c | c c | c c | c c | }
\hline
 & \multicolumn{2}{c|}{VRP20, Cap30} & \multicolumn{2}{c|}{VRP50, Cap40} & \multicolumn{2}{c|}{VRP100, Cap50}  \\
Method & Len  & Gap & Len & Gap & Len & Gap\\
\hline
Gurobi          & 6.10 & 0.00\% &    -      &      -        &       -    &      -     \\
LKH3           & 6.14 & 0.58\% & 10.38 & 0.00\%  & 15.65 & 0.00\% \\
\hline
RL (greedy)\cite{c2}   & 6.59 & 8.03\% & 11.39 & 9.78\%  & 17.23 & 10.12\% \\
AM (greedy)\cite{c5}  & 6.40 & 4.97\% & 10.98 & 5.86\% & 16.80 & 7.34\%\\
\hline
AM-D (greedy) & \textbf{6.28} & \textbf{2.95\%} & \textbf{10.78} & \textbf{3.85\%} & \textbf{16.40} & \textbf{4.79\%}\\
AM-D (2OPT)  &   6.25  &  2.46\%  & 10.73  & 3.37\%   &  16.27  &  3.96\%\\
\hline
\hline
AM-D ($n=20$)      & - & - & 11.00 & 5.97\% & 17.37 & 10.99\%\\
AM-D ($n=50$)      & 6.48 & 6.23\% & - & - & 16.55 & 5.75\% \\
AM-D ($n=100$)    & 6.65 & 9.02\% & 11.04 & 6.36\% & - & - \\
\hline
\end{tabular}\label{t1}
\end{table}

\begin{table}[b]

\caption{Training time and testing time of AM-D. Testing time is average runtime on test instance.}
\centering
\renewcommand\arraystretch{1.25}
\scalebox{1}{
\begin{tabular}{| c  | p{0.15\linewidth}  | p{0.15\linewidth}  | p{0.15\linewidth} | }
\hline
            &  \multicolumn{1}{c|}{VRP20} & \multicolumn{1}{c|}{VRP50} &  \multicolumn{1}{c|}{VRP100}\\
\hline
Training time   &  \multicolumn{1}{c|}{14h} &  \multicolumn{1}{c|}{58h} &  \multicolumn{1}{c|}{250h}\\
\hline
Testing time (greedy)  & \multicolumn{1}{c|}{0.29ms} & \multicolumn{1}{c|}{2.51ms} & \multicolumn{1}{c|}{15.92ms}\\
\hline
Testing time (2OPT)  &  \multicolumn{1}{c|}{0.05s}  & \multicolumn{1}{c|}{0.34s}  &  \multicolumn{1}{c|}{2.21s}\\
\hline
\end{tabular}\label{t3}
}
\end{table}

\subsection{Results and Discussions}

\subsubsection{Comparison Results}
TABLE \ref{t1} shows the results of VRP. Compared with AM, the performance of AM-D is notably improved for both VRP20 (2.02\%), VRP50 (2.01\%) and VRP100 (2.55\%). AM-D significantly outperforms other baseline models as well.

The numerical experiments indicate that AM-D performs better than AM and other baseline methods. AM-D introduces a dynamic encoder-decoder architecture to explore structure features dynamically and exploit hidden structure information effectively at different construction steps. Hence, more hidden and useful structure information is taken into account, thereby leading to a better solution.

\subsubsection{Generalization to Larger or Smaller Instances}
%In order to investigate the generalization performance of our model, experiments are conducted on unseen instances. The results are shown at the last three rows in TABLE \ref{t1} and TABLE \ref{t2}. Specifically, for VRP as shown in TABLE \ref{t1}, the models trained with instances with 20, 50 and 100 customer nodes are named AM-D ($n=20$), AM-D ($n=50$) and AM-D ($n=100$), respectively. All these trained models are generalized to solve unseen instances. For example, the model, AM-D ($n=20$), are used to solve unseen instances with 50 and 100 customer nodes.

How does the performance of the learned heuristics generalize to test instances with larger or smaller customer node size? Experiments are conducted to investigate the generalization performance of AM-D. Specifically, the model trained with instances with 20, 50 and 100 customer nodes are denoted as AM-D ($n=20$), AM-D ($n=50$) and AM-D ($n=100$), respectively. AM-D ($n=20$) is tested on instances with 50 and 100 customer nodes,  AM-D ($n=50$) is tested on instances with 20 and 100 customer nodes, and AM-D ($n=100$) is tested on instances with 20 and 50 customer nodes, respectively.

The results are shown at the last three rows in TABLE \ref{t1}. Specifically, on the one hand, the model trained with small instances ($n=20$) has a good performance on large instances ($n=50$, $n=100$), and the results even better than some baseline methods. On the other hand, the model trained with large instance ($n=100$) performs good on small instance ($n=20$, $n=50$) as well. The reason why AM-D has a good generalization performance may be as follows. AM-D constructs the solution incrementally, and this process can be divided into many stages. At each stage, only a partial solution is constructed, and thus the instance is transformed to a smaller one which is easier to solve.

\subsubsection{Combination With Local Search}
Local search is applied to further improve the results as in \cite{c4}. Firstly, for each instance, a solution is constructed by AM-D (greedy), then the 2OPT local search algorithm is applied to improve this solution. The resultant method is named AM-D (2OPT) and the results are shown in TABLE \ref{t1}. The runtime of AM-D (greedy) and AM-D (2OPT) are also given in TABLE \ref{t3}. The results indicate that the quality of the solution is improved by integrating local search, but the local search brings additional computational cost.

%Though the latter is obviously slower than the former, the extra runtime of AM-D (2OPT) is acceptable in some scenarios that pay more attention to optimality.

\subsubsection{Discussions}
%Machine learning based algorithm vs. operations research algorithm
Machine learning and optimization are closely related, machine learning is often used as an assistant or helper component to improve the performance of solution or reduce computational costs in many optimization algorithms \cite{c32}. Totally different from these methods, AM-D is aiming to learn heuristics from data directly without human intervention. It means that knowledge or features can be extracted from the given problem instances automatically. Specifically, given an optimization problem and its instances generated from distribution $\mathcal{S}$, AM-D can learn an approximation or heuristic algorithm and solve the problem on unseen instances generated from distribution $\mathcal{S}$.

AM-D can be divided into training and testing phases like most of machine learning algorithms. The elapsed time of training and testing are shown in TABLE \ref{t3}. Though the process of training is time-consuming, it is upfront, offline computation and can be seen as searching in algorithm space. Then, the trained model can be used directly to solve unseen instances without retraining from scratch, which is online even real-time computation process. Taking VRP20 for example, it takes 14 hours in training phase, but the process is one-time. Once the model has been trained, it only spends 0.29 milliseconds for solving each instance without retraining. Thus, AM-D is different from the classic heuristics, which search the solution iteratively in solution space from scratch for each instance.

The training phase of AM-D is time-consuming, thus it is trained only for problem instances with small and medium size due to the limitation of computing resources. It is promising to adopt existing parallel computing techniques to improve the computational efficiency for scaling to larger problem instances in the future.

%Recently, \cite{c35} trys to solve large-scale instance for 0-1 based combinatorial optimization problems.
\section{Conclusion}\label{s8}
This paper presents a dynamic attention model with dynamic encoder-decoder architecture for VRP. The key improvement is that the structure features of instances are explored dynamically, and hidden structure information is exploited effectively at different construction steps. Hence, more hidden and useful structure information is taken into account, for constructing a better solution. AM-D is tested by a challenging NP-hard problem, VRP. The results show that the performance of AM-D is better than AM and other baseline models for both problems. In addition, AM-D also shows a good generalization performance on different problem scales.

In the future, the proposed learning heuristic based method, AM-D, can be extended to solve some real-world complex VRP variants \cite{c24, c25, c26, c27, c28, c29} by hybridizing with operations research method, such as VRP with time windows, which will open a new era for combinatorial optimization algorithms \cite{c30}.

\section*{Acknowledgment}
This work is supported by the National Key R\&D Program of China \\
(2018AAA0101203), and the National Natural Science Foundation of China (61673403, U1611262).

%
% ---- Bibliography ----
%
% BibTeX users should specify bibliography style 'splncs04'.
% References will then be sorted and formatted in the correct style.
%
\bibliographystyle{splncs}

\bibliography{ref}

\end{document}